%% file: main.tex
\documentclass{article}
\usepackage{spconf,amsmath,graphicx}
\usepackage{multirow}
\usepackage{float}
\usepackage{algorithm}
\usepackage{algpseudocode}
\usepackage[fontsize=9pt]{fontsize}
\usepackage{bold-extra}
\usepackage[T1]{fontenc}
\usepackage[subtle]{savetrees}
\usepackage{amssymb}
\usepackage{pifont}
\usepackage{url}
\usepackage{hyperref}
\usepackage{ctable}
\usepackage{cleveref}
\usepackage{booktabs}

\newcommand{\comment}[1]{}

\newcommand{\Tref}[1]{Table~\ref{#1}}

\newcommand{\Fref}[1]{Figure~\ref{#1}}

\newcommand{\Sref}[1]{Section~\ref{#1}}

\newcommand{\app}{\raise.17ex\hbox{$\scriptstyle\sim$}}

\usepackage{amsmath}
\DeclareMathOperator*{\argmax}{argmax} 

\usepackage{colortbl,xcolor}
\usepackage{tabularx}
\usepackage{makecell}

\usepackage{xspace}
\makeatletter
\def\onedot{\ifx\@let@token.\else.\null\fi\xspace}
\makeatother

\def\etal{\textit{et al}\onedot}

\usepackage{eucal}

\title{Hindi as a second language:\\ Improving visually grounded speech with semantically similar samples}
\name{Hyeonggon Ryu*, Arda Senocak*, In So Kweon, Joon Son Chung \thanks{*These authors contributed equally to this work. This work was supported by the Institute of Information \& communications Technology Planning \& Evaluation (IITP) grant funded by the Korean government (MSIT) (No. 2022-0-00064, Development of Human Digital Twin Technologies for Prediction and Management of Emotion Workers’ Mental Health Risks)}}
\address{
Korea Advanced Institute of Science and Technology, South Korea}

\begin{document}

%
\maketitle
\begin{abstract}
The objective of this work is to explore the learning of visually grounded speech models (VGS) from multilingual perspective. Bilingual VGS models are generally trained with an equal number of spoken captions from both languages. However, in reality, there can be an imbalance among the languages for the available spoken captions. Our key contribution in this work is to leverage the power of a high-resource language in a bilingual visually grounded speech model to improve the performance of a low-resource language. We introduce two methods to distill the knowledge of high-resource language into low-resource languages: (1) incorporating a strong pre-trained high-resource language encoder and (2) using semantically similar spoken captions. Our experiments show that combining these two approaches effectively enables the low-resource language to surpass the performances of monolingual and bilingual counterparts for cross-modal retrieval tasks.
\end{abstract}

\input{01_intro.tex}
\input{02_approach.tex}

\input{03_experiments.tex}

\input{04_conclusion.tex}

\newpage
\bibliographystyle{IEEEbib}
\bibliography{shortstrings,refs}

\end{document}

%% file: 01_intro.tex
\vspace{-2mm}\section{Introduction}\label{sec:intro}
Visually grounded speech (VGS) models aim to leverage visual information and speech audio without textual information. A growing body of works~\cite{harwath2015deep,Sun16LookListenDecode,harwath2018jointly,Chrupaa2017RepresentationsOL,Kamper2017VisuallyGL,kamper2018visually,kamper2019semantic,rouditchenko2020avlnet,Higy2020TextualSF,PengWordDiscovery22} are introduced to obtain semantic information from the speech by aligning the paired speech and the visual information in a shared embedding space. These models are capable of associating the spoken words with visual instances without any transcription or annotation; such as the spoken word ``lion'' and the images that contain lion objects. Initially, these approaches are investigated in monolingual settings~\cite{harwath2017learning,harwath2018jointly,Chrupaa2019SymbolicIB,Harwath2020Learning,Olaleye2021AttentionBasedKL} where the images are paired with spoken captions from one language. The speech $\leftrightarrow$ image (or vice-versa) retrieval tasks or cross-modal localization tasks are performed to measure the ability of the learned models. 

While the language we use to describe the visual world around us may change, the visual concepts share a similar appearance universally. Since the natural correspondence between monolingual spoken captions and images is learned in earlier studies, the images can be used as a bridge to find correspondences between the different languages. To this end, bilingual models are introduced by using visual information as a transitive relationship between the two languages~\cite{Kamper2018VisuallyGC,harwath2018vision,fei2021cross,RouditchenkoCas,SurisEV22}. These models demonstrate that jointly learned models with spoken captions in two languages perform better than their monolingual counterparts. Additionally, using this cross-lingual correspondence enables one to perform cross-lingual audio-to-audio retrieval or cross-lingual image description tasks. Furthermore, these models are extended by using more languages~\cite{ohishi2020trilingual}.

\vspace{-2mm}

\begin{figure}[ht!]
\centering
{
\resizebox{\linewidth}{!}{
\begin{tabular}{c}
\includegraphics[width = 1\linewidth]{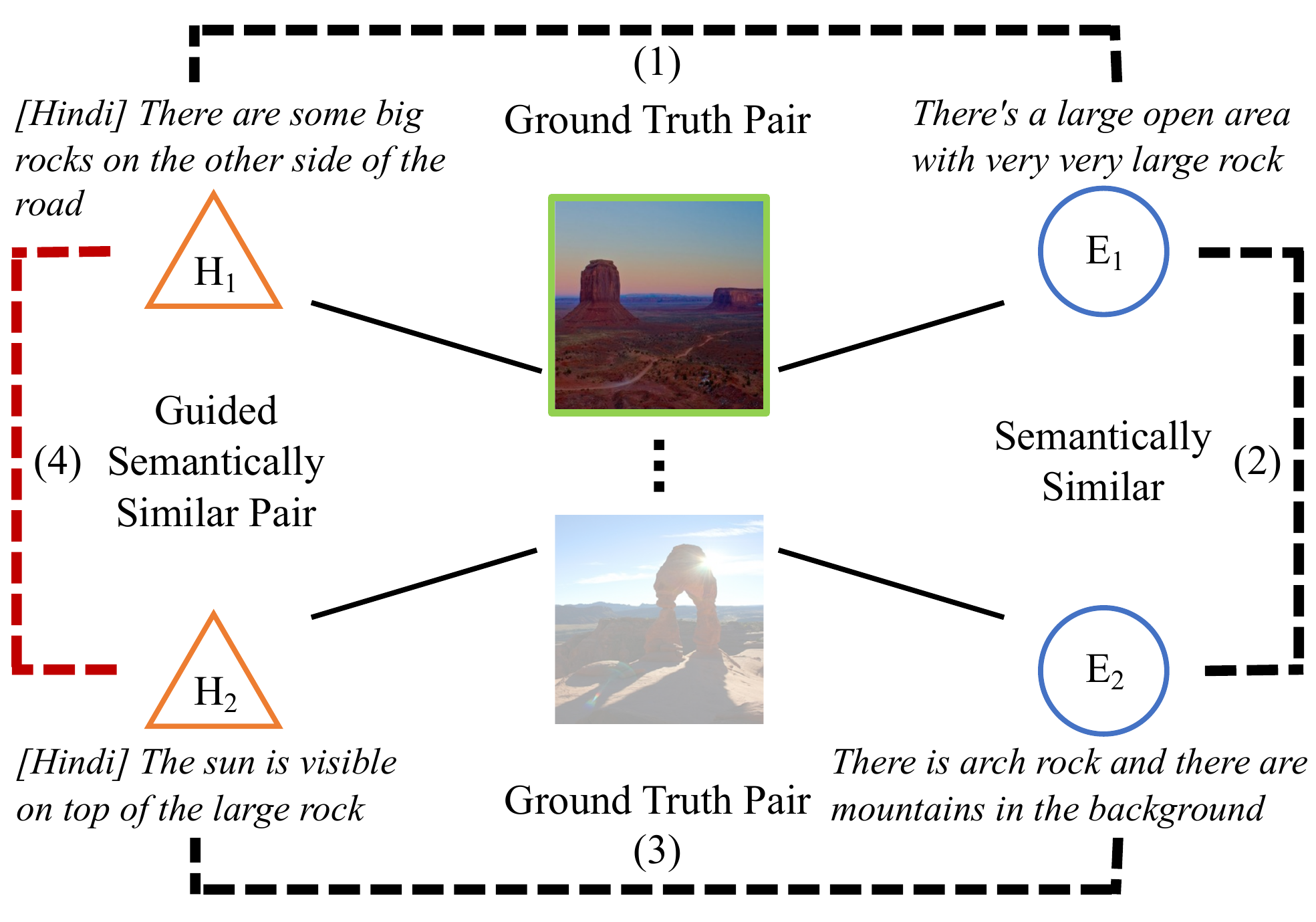} \\
\end{tabular}
}
}\vspace{-2mm}
\caption{\textbf{Semantically similar expressions in VGS.} We can describe a visual scene (green frame) with similar but not necessarily identical
expressions in a language (E1 and E2). By using these similar expressions and the ground truth cross-lingual pairs -- (H1, E1) and (H2, E2) -- in bilingual VGS datasets, we can find highly related samples in another language (H1, H2), and add these intra-language links to the loss.}
\label{fig:teaser}
\vspace{-2mm}
\end{figure}

Prior multilingual visually grounded speech models~\cite{harwath2018vision,ohishi2020trilingual} use an equal number of samples for cross-languages due to the corresponding triplet pairs of (L1, Image, L2) - L stands for language - during training. For example,~\cite{harwath2018vision} uses 100K samples from both English and Hindi to jointly train the model.
However, in reality, there may be more available data for one of the languages for VGS-based learning.  In other words, the quantity of the captions is imbalanced among the languages. Thus, we approach the problem of learning bilingual visually grounded speech models from a different perspective. We assign these cross-languages as high-resource source language and low-resource target language based on the availability of the data size for training in the context of VGS. We pose the question ``Can a high-resource language provide help to a low-resource language?''. In this work, we investigate leveraging the power of a richer language in a bilingual visually grounded speech model to improve the performance of a low-resource language. We demonstrate a proof-of-concept setting where English is used as a high-resource language, and Hindi and Japanese are used as low-resource languages. 

\begin{figure*}[h]
\centering
{
\resizebox{0.9\linewidth}{!}{
\begin{tabular}{c}
\includegraphics[width = 1\linewidth]{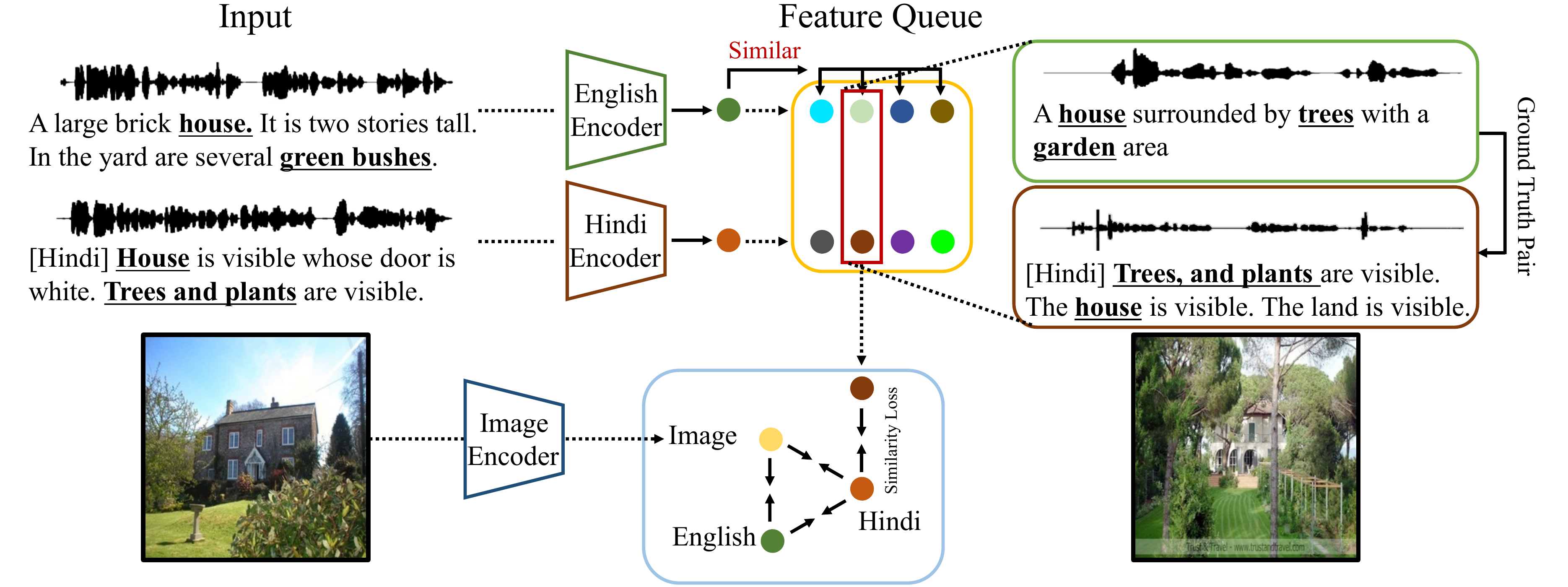} \\
\end{tabular}
}
}

\caption{\textbf{Our proposed bilingual VGS framework.} Our model takes a collection of image-spoken caption triplets. Then each input is encoded in the corresponding encoders. Based on the spoken language representations, semantically similar expressions are obtained from the feature queue. Later, these semantically similar captions are incorporated into bilingual VGS objective as an additional similarity loss to better align the features.}
\label{fig:main}
\vspace{-4mm}
\end{figure*}

To accomplish the task, we first schedule the learning process as; 1) training a high-resource language model in the context of the visually grounded speech, 2) attaching a cross-lingual low-resource language encoder on top of the setup (1) and training it by distilling the knowledge from the strong language model. We then exploit the semantically similar captions of the high-resource language as a transitive relationship to guide the low-resource language while distilling the knowledge. The usage of semantically similar samples embodied from the observation that we can describe a visual scene with similar but not necessarily identical expressions in a language. For simplicity of the example, let’s consider two expressions. The top image in~\Fref{fig:teaser} can be described with both $E_1$ and $E_2$ spoken captions as they are contextually similar. Considering the bilingual VGS datasets, cross-lingual caption pairs are available for each image, such as ($E_1$, $H_1$) and ($E_2$, $H_2$). With the fact that ($E_1$, $H_1$) and ($E_2$, $H_2$) are correspondent pairs, and also ($E_1$, $E_2$) are semantically similar, then by the transitivity we can assume that the remaining expressions, $H_1$ and $H_2$, should be also similar to each other. This forms a cyclic loop as shown in~\Fref{fig:teaser}, which allows learning better-aligned representations for bilingual VGS. With the guidance of the high-resource language’s semantically similar caption, $E_2$, we can find a semantically similar caption in the cross-language, $H_2$, and impose a similarity loss for intra-language expressions in the low-resource language, ($H_1$ $\leftrightarrow$ $H_2$). We note that all of the learning steps listed above are self-supervised. This framework performs favorably to the existing approaches and the baselines on cross-modal (speech $\leftrightarrow$ image) retrieval tasks~\cite{harwath2018jointly,harwath2018vision,ohishi2020trilingual}.

Our main contributions can be summarized as follows: 1) We leverage the power of a rich-resource language in the context of VGS to boost the performance of a low-resource language. 2) We make a new observation of using semantically similar samples as a transitivity relationship to connect cross-languages. 3) We demonstrate that by the guidance of the high-resource language and the semantically similar samples, a low-resource language outperforms its monolingual and bilingual counterparts. 

%% file: 02_approach.tex
\vspace{-4mm}
\section{Approach}\label{sec:approach}
\subsection{Preliminaries}\label{sec:preliminaries}

Our framework takes a collection of image-spoken caption triplets, $X_i=\{v_i, a_{L_{1i}}, a_{L_{2i}}\}$, where $X_i$ is the $i^{th}$ sample in the collection, $v_i$ is an image, $a_{L_{1i}}$ and $a_{L_{2i}}$ are the spoken captions from two different languages. These captions are describing the paired image. Given each image-spoken caption triplet, our backbone networks -- image encoder $f_v(\cdot; \theta_v)$, and the two different spoken caption encoders $f_{L_1}(\cdot; \theta_{a_{L_1}})$ and $f_{L_2}(\cdot; \theta_{a_{L_2}})$ -- encode these triplets into the representations:

\begin{equation}
\begin{aligned}
V_i &= f_v(v_i; \theta_{v_i}), \quad V_i\in\mathbb{R}^{c}\\
A_{L_{1i}} &= f_{L_1}(a_{L_{1i}}; \theta_{a_{L_{1i}}}), \quad A_{L_{1i}}\in\mathbb{R}^{c}\\
A_{L_{2i}} &= f_{L_2}(a_{L_{2i}}; \theta_{a_{L_{2i}}}), \quad A_{L_{2i}}\in\mathbb{R}^{c}.
\end{aligned}
\end{equation}

\subsection{Training}\label{sec:training}

We design our model in the context of bilingual VGS~\cite{harwath2018vision} with the objective of aligning the spoken caption and image embeddings. Thereby, cross-lingual correspondences can be learned via image transitivity. Our entire training procedure is self-supervised. To align the spoken captions and image features in the audio-visual embedding space, representations should be close to each other. A simple approach to enable this is enlarging the similarity of the elements in the image-spoken caption triplets. Similar to~\cite{harwath2018vision,ohishi2020trilingual}, the inner product between the image and each spoken caption features, as well as between the two spoken captions, are computed as a similarity score. We incorporate a batch-wise contrastive learning approach to enforce similarities between image-spoken caption triplet elements. In this manner, we maximize the similarity between each element of the triplet as positive while minimizing the similarity of the pairs that are randomly selected among the samples within the batch as negative. We use InfoNCE~\cite{infoNCE} loss as our contrastive loss function, and it is defined as:

\begin{equation}
L(Z_1,Z_2) = -\frac{1}{N}\sum_{i=1}^N
\left[ \log \frac{\exp(z_{1i}\cdot z_{2i}/\tau)}{\sum_{k=1}^N\exp(z_{1i}\cdot z_{2k}/\tau)} \right],\\
\end{equation}

\noindent where $Z_1 = \{ z_{1i} \}_{i \in [1:N]}$ and $Z_2 = \{ z_{2i} \}_{i \in [1:N]}$ can be a set of image or spoken caption features as we pair the elements in the image-spoken caption triplets. Our learning objective in a bilingual VGS setting is as follows:
\begin{equation}
\mathcal{L}_{base} = L(A_{L_1},V)+L(A_{L_2},V) + L(A_{L_1},A_{L_2}).
\end{equation}

\subsection{Semantically Similar Samples}\label{sec:NN}
As aforementioned, the semantically similar captions can make an alternative bridge in-between the two languages together with the image. Through these connections of semantically similar samples, another similarity loss can be imposed for the intra-language expressions. In this section, we describe the process of selecting semantically similar samples and the additional similarity loss as~\cite{khosla2020superv,HanCoclr,LiDeclip22}. Similar to~\cite{DwibediATSZ21, LiDeclip22, SenocakLearning22}, we use a feature queue $Q$ to select similar samples. This queue holds the spoken caption features extracted during the previous iterations, and it updates the spoken caption features periodically by adding or removing new samples. As cross-lingual spoken captions are paired, they are stored as pairs in the queue as well. During semantically similar expression selection, the input spoken caption features, $z=\{z_1, z_2\}$, are used as a query. Let’s take Language 1 as an example but note that this process is the same for the other language as well. The most similar element in the queue to the input feature $z_1$ is $q_1$. As spoken caption features are stored as pairs in the queue, $\{q_1,q_2\}$ (since they are paired in the dataset), we obtain $q_1$’s cross-lingual correspondence $q_2$ which is the semantically similar expression of the spoken caption feature in Language 2, $z_2$. After selecting $q_2$, an additional similarity loss, $\mathcal{L}_{NN}$, is imposed in between the ($z_2$ $\leftrightarrow$ $q_2$). Please see~\Fref{fig:teaser} and~\Fref{fig:main}. With this loss, similar features in the same language stay close-by in the joint embedding space.

\begin{equation}
\begin{aligned}
P_Q(q_1)&=q_2 \quad where \quad (q_1,q_2) \in Q \\
NN(z_i,Q)&=P_Q(\argmax_{(q_1,q_2) \in Q}{(z_{i1} \cdot q_1)})
\end{aligned}
\end{equation}

\vspace{-4mm}
\begin{equation}
    \begin{aligned}
\mathcal{L}_{NN} = -\frac{1}{N}\sum_{i=1}^N \left[ \log \frac{\exp(NN(z_i,Q)\cdot z_{2i}/\tau)}{\sum_{k=1}^N\exp(NN(z_i,Q)\cdot z_{2k}/\tau)} \right].
    \end{aligned}
\end{equation}

\noindent Finally, we sum up the loss of the baseline bilingual model and the similarity loss from semantically similar samples:

\begin{equation}
\mathcal{L} = \mathcal{L}_{base}+\mathcal{L}_{NN}.    
\end{equation}

%% file: 03_experiments.tex
\begin{table*}[ht!]
\footnotesize
  \caption{\textbf{Summary of retrieval recall scores for all models and baselines.}}
  \label{tab:retrieval_results}
  \centering
  \setlength{\tabcolsep}{5pt}
  \begin{tabular}{l c ccc c ccc c ccc c ccc}
    \toprule
    \multicolumn{1}{c}{} &  \multicolumn{3}{c}{\textbf{H $\rightarrow$ I}} & \multicolumn{3}{c}{\textbf{I $\rightarrow$ H}} & \multicolumn{3}{c}{\textbf{J $\rightarrow$ I}} & \multicolumn{3}{c}{\textbf{I $\rightarrow$ J}} \\
    \cmidrule(lr){2-4}\cmidrule(lr){5-7}\cmidrule(lr){8-10}\cmidrule(lr){11-13}\cmidrule(lr){14-16} 
    \textbf{Model} & \textbf{R@1} & \textbf{R@5} & \textbf{R@10} & \textbf{R@1} & \textbf{R@5} & \textbf{R@10} & \textbf{R@1} & \textbf{R@5} & \textbf{R@10} & \textbf{R@1} & \textbf{R@5} & \textbf{R@10}\\

    \bottomrule
    Harwath~\etal\cite{harwath2018vision} & .080 & .25 & .356 & .074 & .235 & .354 & - & - & - & - & - & -\\
    Ohishi~\etal\cite{ohishi2020pair} & .094 & .298 & .418 & .093 & .295 & .382 & .201 & .497 & .639 & .167 & .443 & .578\\
    Ohishi~\etal\cite{ohishi2020trilingual} & .103 & .299 & .429 & .110 & .295 & .399 & .210 & .515 & .667 & .158 & .435 & .604\\
    Ohishi~\etal\cite{ohishi2020trilingual} +Self Attention & .112 & .315 & .445 & .108 & .313 & .419 & .203 & .520 & .667 & .200 & .468 & .623\\
    \cmidrule(lr){1-16}
    Baseline Monolingual & .081 & .268 & .387 & .09 & .283 & .386 & .144 & .413 & .599 & .179 & .454 & .581\\
    Baseline Bilingual & .098 & .29 & .414 & .125 & .321 & .429 & .149 & .428 & .604 & .182 & .47 & .602 \\
    \cmidrule(lr){1-16}
    Baseline Bi. + NN & .086 & .291 & .413 & .105 & .317 & .429 & .168 & .428 & .582 & .191 & .477 & .622 \\

    Baseline Bi. + Pretrained HRL & .106 & .308 & .442 & .114 & .342 & .444 & .197 & .504 & .648 & .207 & .494 & .628 \\
    \cmidrule(lr){1-16}
    \rowcolor{lightgray!25}
    \textbf{Ours} & \textbf{.132} & \textbf{.343} & \textbf{.471} & \textbf{.144} & \textbf{.354} & \textbf{.462} & \textbf{.224} & \textbf{.536} & \textbf{.677} & \textbf{.211} & \textbf{.517} & \textbf{.664} \\
    \bottomrule
  \end{tabular}\vspace{-4mm}
\end{table*}

\vspace{-2mm}\section{Experiments}\label{sec:experiments}
\subsection{Datasets and Evaluation Metrics}\label{sec:dataset}
In our experiments, we use 1) the PlacesAudio dataset~\cite{harwath2016unsupervised,harwath2017learning}, 2) Hindi spoken captions for a subset of the PlacesAudio~\cite{harwath2018vision}, and 3) Japanese spoken captions for a subset of the PlacesAudio~\cite{ohishi2020trilingual}. The PlacesAudio contains 400K images and English spoken caption pairs where these captions describe their paired images. Hindi spoken captions dataset has 100K images from the Places dataset together with the Hindi descriptions. Similarly, Japanese spoken captions dataset has 100K images and spoken caption pairs from the Places dataset. Note that these datasets share common images that make image-spoken caption triplets. While we use the 400K PlacesAudio dataset to pre-train the high-resource language, the Hindi and Japanese spoken captions dataset are used in bilingual VGS training. The model is evaluated on the 1000 triplets of a validation set with cross-modal retrieval performance. To assess the retrieval performance, Recall at K (R@K) is adopted as the main evaluation metric.

\vspace{-2mm}\subsection{Implementation Details}\label{sec:implementation}
Following Harwath~\etal~\cite{harwath2018vision}, the input image is resized into 256 x 256 pixel, and then randomly cropped as 224 x 224 pixels. Each wavefrom is converted to spectrogram by a series of 25ms frames with a 10ms hop size. 40 log mel filterbank energies are applied to get the mel-spectrogram. The resulting mel-spectrograms are in the size of 1024 x 40 and 3072 x 40 for Hindi and Japanese, respectively. We use spoken caption encoder from~\cite{harwath2018vision} and image encoder is ResNet-18. The output features of all the encoders are presented in 1024 dimensions. The size of the feature queue is 1024. We set $\tau$ as 0.3 in the experiments.

\vspace{-2mm}\subsection{Baselines}\label{sec:baselines}
We introduce the general pipeline of our bilingual VGS architecture in~\Sref{sec:approach}. The baseline models/approaches below are obtained by using different components of the general model based on the experimental setups. We note that for the objective of this work, we rename the spoken caption encoders as high-resource and low-resource language encoders for the remainder of the paper.

\noindent\textbf{Harwath~\etal\cite{harwath2018vision}:} It is a standard baseline for bilingual VGS model in Hindi and English. This model is trained jointly in a self-supervised way by using English, Image, and Hindi language triplets. Margin ranking-based triplet loss is used as an objective function. An equal number of training data, around \app 100K, is used to train both languages.

\noindent\textbf{Ohishi~\etal\cite{ohishi2020pair}:} This prior bilingual VGS method is similar to that of Harwath~\etal\cite{harwath2018vision}, but it uses disjoint datasets. It includes visually similar images (to the anchor image) and their corresponding spoken captions as additional pairs (pair expansion) in the loss function. An equal amount of data is incorporated to train the bilingual models. We use this method as a baseline in an aligned dataset setting.

\noindent\textbf{Ohishi~\etal\cite{ohishi2020trilingual}:} This approach proposes a trilingual VGS model with English, Hindi, and Japanese languages. Similar to Harwath~\etal\cite{harwath2018vision}, this model is also trained jointly with three languages and common images. Though network architecture of this approach is identical to~\cite{harwath2018vision}, additionally a self-attention layer is introduced. Furthermore, masked margin softmax loss is replaced with the triplet loss. Similar to~\cite{harwath2018vision}, this model also uses an equal number of training data to train three languages. 

\noindent\textbf{Baseline Monolingual:} To understand the gap between monolingual, multilingual, and knowledge-distilled bilingual (as we propose) VGS approaches, we train our architecture with only image and a low-resource language encoders in a self-supervised way. 

\noindent\textbf{Baseline Bilingual:} In this setup, another language encoder is attached to the baseline monolingual architecture. Joint self-supervised learning is applied among three encoders as in~\cite{harwath2018vision}. Both language encoders are trained with the same amount of data. Note that this is the default architecture for our model as explained in~\Sref{sec:approach} without using semantically similar samples, only with $\mathcal{L}_{base}$. In terms of architecture, it is similar to~\cite{harwath2018vision}. However, it should be noted that this is simply an updated version of~\cite{harwath2018vision} with recent approaches such as changing the VGG-16 vision encoder to ResNet-18, and triplet loss to InfoNCE loss.

\noindent\textbf{Baseline Bilingual + Pre-trained High-Resource Language:} To quantify the impact of using a pre-trained high-resource language in bilingual VGS models, we train a version of the Baseline Bilingual model where one of the language encoders, a resource-rich language encoder, is pre-trained and fixed (parameters are not getting updated). This pre-trained model is trained in the baseline monolingual setup with a large amount of data, 400K English spoken captions. The other spoken caption encoder, low-resource language, is trained from scratch with the help of this well-trained high-resource language encoder in the Baseline Bilingual model setup. This setup can be considered as knowledge distillation of the rich-resource language into low-resource language in the context of VGS.

\noindent\textbf{Baseline Bilingual + NN:} We train our Baseline Bilingual model with aforementioned semantically similar samples, NN, to see the effect of this approach. We highlight that semantically similar samples are used bidirectionally since each language model is trained jointly. 

\noindent\textbf{Ours:} This is our proposed bilingual model where both pre-trained high-resource language (PHRL) knowledge distillation and NN approaches are used to learn better low-resource language (LRL) models in the context of VGS (~\Fref{fig:main}). We note that the NN approach only follows the guidance of the high-resource language because NN samples in the direction of (LRL $\rightarrow$ PHRL) have no effect as the high-resource language model is pre-trained and fixed (parameters are not getting updated).
\vspace{-4mm}\subsection{Quantitative Results}\label{sec:quantitative}
In this section, we compare our method with the existing VGS and baseline approaches that are introduced in~\Sref{sec:baselines}.

Specifically, we provide the results in a setting where Hindi and Japanese are used as low-resource languages.
Retrieval recall scores for each baseline scenario are shown in~\Tref{tab:retrieval_results}. Key findings are as follows:

\noindent\textbf{1. Training VGS models with multiple languages improve the performance of the target language.} As previously shown in~\cite{harwath2018vision,ohishi2020trilingual}, training our baseline architecture with bilingual setup outperforms the monolingual counterpart both for Hindi and Japanese languages. However, the performance gap is smaller for the Japanese. We hypothesize that this is due to the characteristics of the spoken captions in the dataset as explained in~\cite{ohishi2020trilingual}.

\noindent\textbf{2. The existence of a high-resource language improves the performance of the low-resource language.} We find that using a fixed pre-trained high-resource source language with a large amount of data (a strong resource-rich language) in a bilingual VGS model helps improve performance on low-resource target languages. This shows that leveraging the power of a rich-resource language and distilling the knowledge into low-resource language is an important point to consider when VGS models are designed. 

\noindent\textbf{3. Using semantically similar samples in the absence of a strong language model guidance is not helpful.} As aforementioned, using semantically similar samples can create another pathway to link two languages. However, our experiments show that using semantically similar samples in a bilingual setup does not show improvement. We conjecture that in the absence of strong language models, selected semantically similar samples are not accurate. Thus, they can not provide a proper bridge in-between the two languages. This implies that a resource-rich strong language model is necessary to get an advantage from semantically similar samples.

\noindent\textbf{4. Pre-trained strong high-resource source language and using semantically similar samples (NN approach) are complementary to each other.} The performance of our full model shows that combining a pre-trained resource-rich language model and NN approach significantly boosts the performance of the low-resource target language compared to monolingual or bilingual settings. Additionally, the performance gaps in (Baseline Bi. + Pre-trained HRL vs.\ Ours) and (Baseline Bi. + NN vs.\ Ours) show that usage of NN and pre-trained high-resource source language model are complementary in the context of VGS. It indicates that with the guidance of a strong language, selected semantically similar samples are more accurate. Therefore, these samples link two languages properly and semantically similar samples can be seen as another transitivity in the VGS models.

\noindent\textbf{5. Our full model outperforms standard benchmark methods.} As~\Tref{tab:retrieval_results} displays, our approach outperforms all the prior methods regardless of the language that is used a low-resource language. We highlight that the smaller performance gap in the Japanese language compared to Ohishi~\etal\cite{ohishi2020trilingual} comes from the fact that~\cite{ohishi2020trilingual} uses three languages simultaneously. However, we only use two languages. As it is proven in~\cite{harwath2018vision, ohishi2020trilingual} and in our key finding (1), additional languages bring more performance improvement. Nonetheless, our model surpass a trilingual model. It also demonstrates another important point that our proposed approach compensates for the effect of using additional spoken caption datasets and models. 

\vspace{-4mm}\subsection{Qualitative Results for Semantically Similar Expressions}\label{sec:qualitative}
\begin{figure}[htb!]
\centering
{
\resizebox{0.9\linewidth}{!}{
\begin{tabular}{c}
\includegraphics[width = 1.0\linewidth]{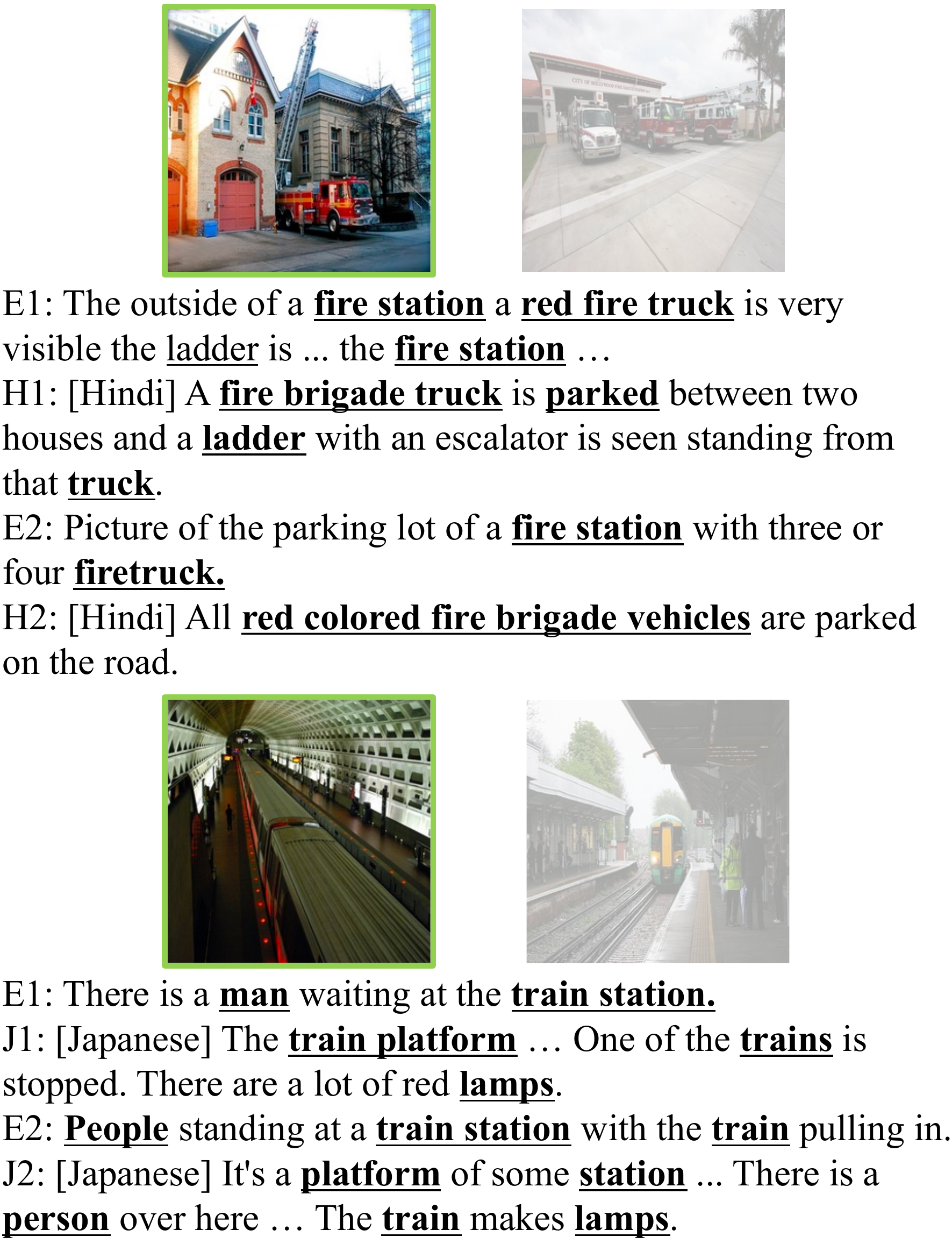} \\
\end{tabular}
}
}\vspace{-2mm}
\caption{\textbf{Obtained semantically similar expressions.}}
\label{fig:qual}
\vspace{-4mm}
\end{figure}

In this section, we visualize some of the selected semantically similar expressions together with the images and paired input spoken captions in our proposed bilingual VGS model. Our qualitative results in~\Fref{fig:qual} show that selected samples indeed describe the images and form the cyclic cross-lingual relationships as described in~\Sref{sec:intro} and~\Fref{fig:teaser}.

%% file: 04_conclusion.tex
\vspace{-2mm}\section{Conclusion}\label{sec:conclusion}
In this paper, we approach the problem of bilingual VGS models from the perspective of imbalanced spoken captions quantity among the languages. We introduce two approaches -- leveraging a strong pretrained source language and using semantically similar samples -- to distill the knowledge of resource rich language to low-resourced language in order to boost the performance. We suggest the learning steps to effectively combine these two approaches as it requires careful design choices. As a result, our method enables a low-resourced language to outperform its monolingual and bilingual counterparts for cross-modal retrieval tasks.